\documentclass{ifacconf}

\usepackage{graphicx}      
\usepackage{natbib}        

\usepackage{amsmath, amssymb}
\usepackage{mathtools}
\usepackage{xurl}
\usepackage{multirow}
\usepackage{subcaption}

\DeclareMathOperator*{\minimize}{minimize}

\renewcommand{\vector}[1]{\boldsymbol{#1}}
\renewcommand{\matrix}[1]{#1}
\newcommand{\transp}{^\top}

\usepackage{xcolor}

\usepackage[absolute,overlay]{textpos}

\begin{document}
\begin{frontmatter}

\begin{textblock*}{\textwidth}(-0.5cm,3cm) 
   This work has been submitted to IFAC for possible publication.
\end{textblock*}

\title{MPC-based Motion Planning for Autonomous Truck-Trailer Maneuvering\thanksref{footnoteinfo}}


\thanks[footnoteinfo]{This work has been carried out within the
framework of projects Flanders Make SBO DIRAC: Deterministic and Inexpensive Realizations of Advanced Control, Flanders Make SBO ARENA: Agile and Reliable Navigation and Flanders Make SBO FLEXMOSYS: Flexible Multi-Domain Design for Mechatronic Systems.}

\author[First]{Mathias Bos\textsuperscript{1}}   
\author[First]{Bastiaan Vandewal\textsuperscript{1}} \thanks{These authors contributed equally.}
\author[First]{Wilm Decr\'e}
\author[First]{Jan Swevers}

\address[First]{MECO Research Team, Department of Mechanical Engineering,\\
KU Leuven, Belgium (e-mail: firstname.lastname@kuleuven.be)\\
and DMMS lab,
Flanders Make,
Leuven, Belgium}

\begin{abstract}                
Time-optimal motion planning of autonomous vehicles in complex environments is a highly researched topic. This paper describes a novel approach to optimize and execute locally feasible trajectories for the maneuvering of a truck-trailer Autonomous Mobile Robot (AMR), by dividing the environment in a sequence or route of freely accessible overlapping corridors. Multi-stage optimal control generates local trajectories through advancing subsets of this route. To cope with the advancing subsets and changing environments, the optimal control problem is solved online with a receding horizon in a Model Predictive Control (MPC) fashion with an improved update strategy. This strategy seamlessly integrates the computationally expensive MPC updates with a low-cost feedback controller for trajectory tracking, for disturbance rejection, and for stabilization of the unstable kinematics of the reversing truck-trailer AMR. This methodology is implemented in a flexible software framework for an effortless transition from offline simulations to deployment of experiments. An experimental setup showcasing the truck-trailer AMR performing two reverse parking maneuvers validates the presented method.
\end{abstract}


\end{frontmatter}
\section{Introduction}
\label{section:introduction}
\subsection{Challenges}
To increase the capability and flexibility of autonomous vehicles in warehouses, greenhouses and factory floors, Autonomous Mobile Robots (AMRs) should be able to perform complex maneuvers autonomously to successfully navigate in obstructed environments. Such maneuvers involve the need to model, formulate, and solve a challenging motion planning problem, and to execute the found solution with an appropriate control strategy.

\begin{figure}
    \centering
    \includegraphics[width=0.8\linewidth]{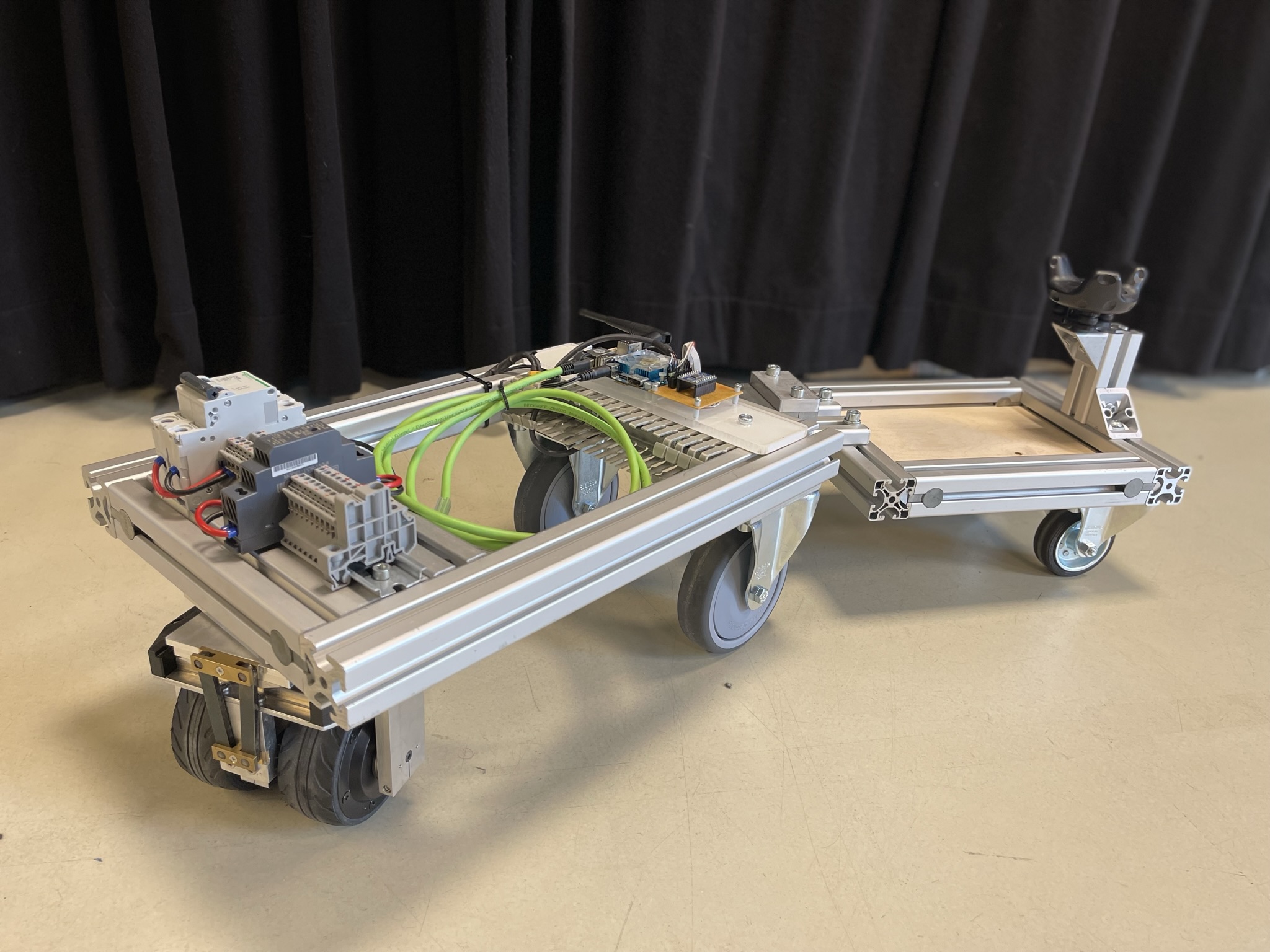}
    \caption{Truck-trailer AMR for lab experiments.}
    \label{fig:truck-trailer_AGV}
\end{figure}

The trajectory optimization problems which define these maneuvers typically use simple vehicle models with a body geometry that does not change over time. In this paper, we consider a more complex vehicle that consists of a truck with a trailer, with corresponding kinematic vehicle model, to perform a backwards parking maneuver. The approach is demonstrated on a small-scale lab setup, displayed in Fig.~\ref{fig:truck-trailer_AGV}, which mimics a real-life vehicle with trailer.

The non-holonomic nature of the vehicle and non-convex kinematic constraints make this parking maneuver in general challenging. Additionally, one needs to extend the kinematic vehicle model with the trailer kinematics and to account for more complex collision avoidance constraints. Finally, while driving backward, the vehicle kinematics are unstable, such that fast, stabilizing control is required. To cope with these complexities, multiple strategies are proposed in literature. 

\subsection{Related Work}
For a set of maneuvers, one of which is a parking scenario, \cite{Ghilardelli} use a parameterized ninth-order polynomial curve (spline) to generate smooth and feasible paths. This method is only appropriate for offline path planning, which limits its practical use. Alternatively, \cite{Evestedt} propose to use a Rapidly Exploring Random Tree (RRT) with motion primitives to generate motion plans that are kinematically feasible and include the limitations with respect to the performance of the stabilizing tracking controller. Two major drawbacks of this method are the need for extensive sampling and, as the authors state, undesired non-intuitive solutions.

Another sampling-based method, with full-scale demonstrations of a truck with a trailer driving backward, is presented in \cite{Ljungqvist2017LatticebasedMP} and \cite{Ljungqvist_real_truck}, where graph-search algorithms are used to select the optimal trajectory using a regular state lattice with a finite set of kinematically feasible motion primitives. This makes the problem tractable for real-time applications but limits the maneuverability of the vehicle due to the discretization of the state space. Generally, frameworks based on sampling methods lack guarantees for completeness and are only asymptotically optimal.

To stabilize the backward motion of the truck with trailer around piecewise linear and circular path segments, \cite{Evestedt} and  \cite{Ljungqvist2017LatticebasedMP} use a cascaded path tracking control approach with a hybrid linear quadratic (LQ) controller. \cite{Ljungqvist_real_truck} improve the approach to be suited for more situations, including segments that switch between forward and backward driving, by deriving a full state feedback controller for which stability is again proven for linear and circular segments.

A detailed survey of truck-trailer kinematics, which discusses some properties of the general n-trailer is presented by \cite{altafini}.  We will only consider a single trailer in this paper.

\subsection{Contributions}
In this paper we present a novel approach to the optimal motion planning and control for the maneuvering of a truck-trailer AMR which consists of two pillars: \\
1) Multi-stage optimal motion planning through convex corridors, which starts from the work of \cite{MercyTim2018O} on optimal navigation through vast environments. The idea behind this work is to subdivide a complex world with many obstacles into a series of convex (in Mercy's case rectangular) frames that fill the free space between static obstacles, through which a local optimizer plans optimal motion trajectories for vehicles with simple geometries and kinematics. This paper extends this work by adding geometric and kinematic complexity. The geometry of the truck and trailer combination changes the approach compared to a traveling point mass as the truck and trailer do not enter or leave a frame at the same time, or in the same order.\\
Optimal control has the advantage of not being restricted to motion primitives, which allows it to be more flexible in finding solutions to complex maneuvering problems, and moreover, it produces optimal trajectories that can be recomputed online with a receding planning horizon.

2) Model Predictive Control (MPC) with an improved update strategy to cope with longer computation times, while additional linear time-invariant feedback control ensures stabilization, accurate tracking and disturbance rejection at a higher rate than at which the optimization problem can be solved given state-of-the-art solvers. As \cite{Neunert2016} describe, the MPC is then responsible for trajectory planning and the rejection of low-frequency disturbances, while the trajectory tracking feedback controller is responsible for the rejection of high-frequency disturbances.

The methodology of the two pillars is implemented in a flexible software framework that enables developers to easily simulate the behavior of their algorithms with a virtual vehicle, and to deploy them on the real experiment setup without additional programming effort.

\subsection{Structure}
Section~\ref{method} describes the multi-stage implementation of the optimal control problem after introducing the considered vehicle model. It also discusses the structure of the stabilizing feedback control. Section~\ref{experiments} presents the experimental work to validate the method and these results are discussed subsequently. Finally, Section~\ref{conclusion} suggests possible extensions and concludes the paper.
\section{Methods}
\label{method}
This section first presents the kinematic vehicle model that is used in the multi-stage optimal motion planning problem. After discussing the motion planning problem formulation, the optimization toolchain, and the strategy to solve it repeatedly, it presents the used feedback controller. It concludes with the structure of the software framework that allows for fast testing and deploying of the methods.

\subsection{Kinematic Vehicle Model}
\label{subsection:model}

\begin{figure}
    \centering
    \includegraphics[width=0.4\linewidth]{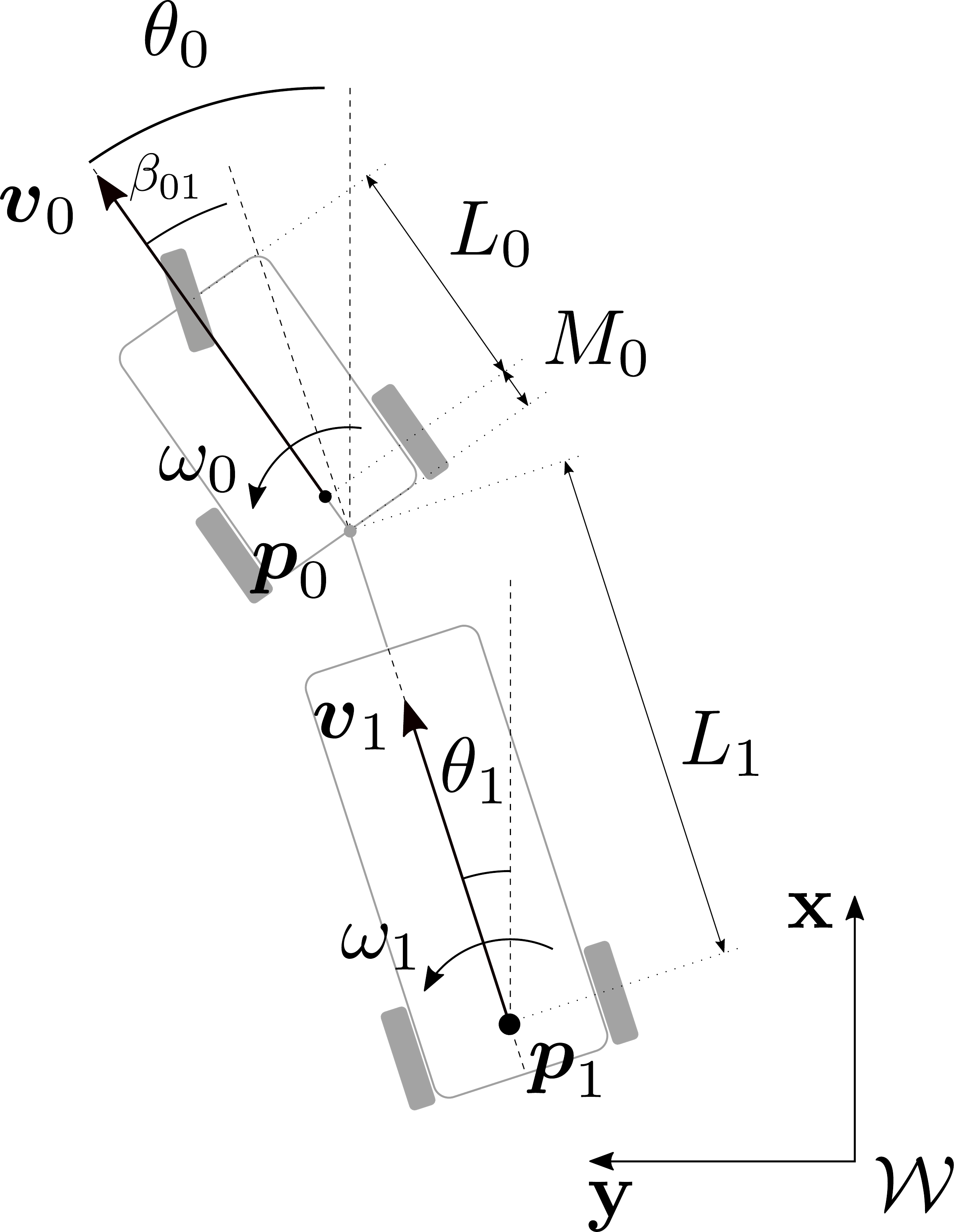}
    \caption{Geometric parameters and variables of a truck-trailer AMR.}
    \label{fig:vehicle_geometry}
\end{figure}

The nonlinear kinematics of a truck-trailer AMR are described by a set of Ordinary Differential Equations (ODE) of the form $\dot{\vector{x}}(t) = \vector{f}(\vector{x}(t), \vector{u}(t))$. Referring to the truck as vehicle~0 and the trailer as vehicle~1, the state of the combined vehicles can be fully represented by the state vector $\vector{x}=[p_{x,1} \: p_{y,1} \: \theta_1 \: \theta_0]\transp \in \mathbb{R}^4$, where $[p_{x,1} \: p_{y,1}]\transp = \vector{p}_1$ is the center point of the trailer axle, $\theta_1$ is the orientation of the trailer and $\theta_0$ is the orientation of the truck with respect to the world frame $\mathcal{W}$. The control input vector $\vector{u} = [v_0 \: \omega_0]\transp$ consists of the longitudinal velocity $v_0$ and the rotational velocity $\omega_0$ of the truck. If we assume zero slip of the driven wheel, the ODE for this vehicle can be written as in \cite{altafini}: 
\begin{equation}
\begin{aligned}
\dot{p}_{x,1}(t)&=v_1(t)\cdot \cos\theta_1(t), \\
\dot{p}_{y,1}(t)&=v_1(t)\cdot \sin\theta_1(t), \\
\dot{\theta}_1(t)&=\frac{v_0(t)}{L_1}\cdot\sin\beta_{01}(t) - \frac{M_0}{L_1}\cdot\cos\beta_{01}(t)\cdot\omega_0(t), \\
\dot{\theta}_0(t)&= \omega_0(t),
\end{aligned}
\end{equation}
where $v_1(t)$ is given by
\begin{equation}
v_1(t) = v_0(t)\cdot \cos\beta_{01}(t) + M_0\cdot \sin\beta_{01}(t)\cdot\omega_0(t),
\end{equation}

where the states, controls, and geometric parameters are as indicated in Fig.~\ref{fig:vehicle_geometry}. $\beta_{01} = \theta_0 - \theta_1$ is the relative angle between the truck and the trailer; $L_i$ is the distance between the steering wheel and the axle center (or, equivalently, the hitching point and the axle center) for vehicle~$i$; $M_i$ is the distance between the axle center and the hitching point of vehicle~$i$.


\subsection{Multi-stage Optimal Motion Planning}
\begin{figure}
    \centering
    \includegraphics[width=0.5\linewidth]{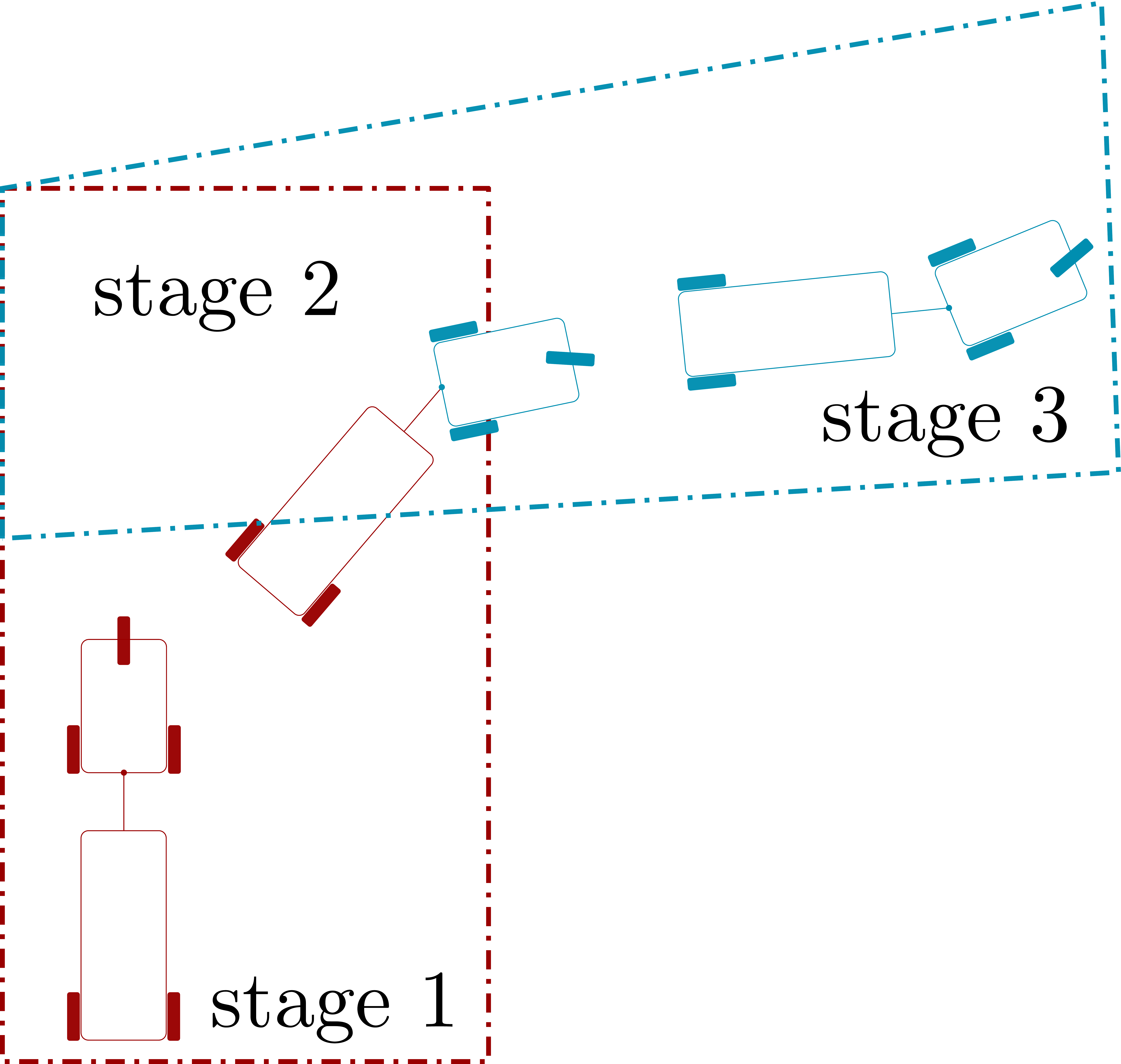}
    \caption{The three stages of the OCP: both vehicles in the first corridor (red); one vehicle in the second and one vehicle in the first corridor (blue and red); both vehicles in the second corridor (blue).}
    \label{fig:stages}
\end{figure}

The motion planning approach is based on a division of free space into convex polyhedrons, referred to as corridors, each with an associated feasible waypoint. The motion planner assumes the availability of a sequence of corridors or route that guides the truck-trailer AMR from the initial to the terminal state. An optimal control problem (OCP) is then formulated as finding an optimal trajectory through two subsequent corridors. The selection of these two corridors shifts along the route in a receding horizon fashion. Their region of overlap is required to be at least large enough to contain each of the truck and trailer vehicles, although not simultaneously. For every dual-corridor combination the OCP is divided into three stages, as shown in Fig.~\ref{fig:stages}. In the first stage, both truck and trailer are fully located in the first corridor. The second stage has one vehicle in the first corridor and the other vehicle in the second corridor, depending on the moving direction of the vehicle. In the third stage both vehicles are in the second corridor. 

This multi-stage approach is detailed for aerial vehicles in \cite{Mathias}, although their method only allows a point mass moving through consecutive corridors. Considering point mass motion simplifies the procedure that is discussed above, as a point mass can instantly only be present in one of the corridors outside the overlap region, whereas the truck-trailer AMR can simultaneously exist in two corridors outside the overlap region.

The vehicle kinematics are discretized using a multiple shooting scheme with fourth order Runge-Kutta integration. These discrete dynamics are represented by $\vector{x}_{k+1} = \vector{F}(\vector{x}_{k},\vector{u}_{k},T)$, with $T$ the variable motion time of the stage. Stitching constraints allow to connect consecutive stages and require full state and control vector equality at the boundary of the stages.

In each of the stages both the truck and the trailer are constrained to be present in one of the corridors. All vertices of both vehicles are constrained using the formulation described by \cite{Mathias}. Each side of the corridor is defined using the standard equation of a line, the parameter vectors of which are combined in the matrix $\matrix{W}$, e.g. $\matrix{W}_{\text{v0}}$ for vehicle~0. The matrix $\matrix{P}_{\text{v0},j,k}$ contains the homogeneous coordinate representation of all vertices of vehicle~0 using state information and vehicle dimensions for time~$k$ in stage~$j$. In order to nudge the vehicle away from the corridor boundaries, some slack variables are combined in a matrix~$\matrix{S}$. These slack variables will be introduced in the objective which tries to keep their values as close as possible to a safety distance~$s_d$ from the corridor wall.

The OCP for the current setup is formally expressed as 
\begin{equation}
  \begin{alignedat}{5}
  \minimize_{\substack{\vector{x}_{j,k}, \vector{u}_{j,k}, T_j, \\ \matrix{S}_{\text{v0},j,k}, \matrix{S}_{\text{v1},j,k}}} &\hspace{0.1em}& \sum_{j=1}^{3} \left(T_j\: +\: \sum_{i=0}^{1}\sum_{k=0}^{N_j}w_i(\matrix{S}_{\text{v}i,j,k}+s_{d})^2\right)\span\span\span \\
  \text{subject to}      &\hspace{0.1em}& \vector{x}_{1,0}   &= \vector{x}_{0}\\
                          &\hspace{0.1em}& \vector{x}_{1,k+1} &=\vector{F}(\vector{x}_{1,k},\vector{u}_{1,k},T_1) &\:& {\scriptstyle \text{for } k \in \left[0,N_1-1\right]}&\\
                          &\hspace{0.1em}& \vector{x}_{1,N_1} &= \vector{x}_{2,0}\\
                          &\hspace{0.1em}& \\
                          &\hspace{0.1em}& \vector{x}_{2,k+1} &=\vector{F}(\vector{x}_{2,k},\vector{u}_{2,k},T_2) &\: & {\scriptstyle \text{for } k \in \left[0,N_2-1\right]}&\\
                          &\hspace{0.1em}& \vector{x}_{2,N_2} &= \vector{x}_{3, 0}\\
                          &\hspace{0.1em}& \\
                          &\hspace{0.1em}& \vector{x}_{3,k+1} &=\vector{F}(\vector{x}_{3,k},\vector{u}_{3,k},T_3) &\:& {\scriptstyle \text{for } k\in \left[0,N_3-1\right]}&\\
                          &\hspace{0.1em}& \vector{x}_{3,N_3}   &= \vector{x}_{f}&\\
                          &\hspace{0.1em}& \\
                          &\hspace{0.1em}& \matrix{W}_{\text{v0},j}\transp \matrix{P}_{\text{v0},j,k} &\leq \matrix{S}_{\text{v0},j,k} \leq 0 &\:&{\scriptstyle \forall j, k}&\\
                          &\hspace{0.1em}& \matrix{W}_{\text{v1},j}\transp \matrix{P}_{\text{v1},j,k} &\leq \matrix{S}_{\text{v1},j,k} \leq 0 &\:&{\scriptstyle \forall j,k}\\
                          &\hspace{0.1em}& \vector{u}_{min} &\leq \vector{u}_{j,k} \leq \vector{u}_{max} &\:&{\scriptstyle \forall j,k}&\\
                          &\hspace{0.1em}& \dot{\vector{u}}_{min} &\leq \dot{\vector{u}}_{j,k} \leq \dot{\vector{u}}_{max} &\:&{\scriptstyle \forall j,k}&\\
                          &\hspace{0.1em}& \beta_{01,min} &\leq \beta_{01,j,k} \leq \beta_{01,max} &\:&{\scriptstyle \forall j,k},
   \end{alignedat}
\end{equation}
where $N_1$, $N_2$ and $N_3$ are the horizon lengths of the three stages and their corresponding stage times are defined as the optimization variables $T_1$, $T_2$ and $T_3$. The initial state and the desired terminal state are denoted by $\vector{x}_0$ and $\vector{x}_f$ respectively. $w_0$ and $w_1$ are weighting factors to express the importance of the allowed safety margin. Both control inputs, combined in the control input vector $\vector{u}$, are piecewise linearly parameterized such that both the inputs and their derivatives can be bounded. An additional constraint on the angle between truck and trailer $\beta_{01}$ is added to account for the physical shape of the AMR.

Since the stage timings $T_1$, $T_2$ and $T_3$ are free optimization variables, it is plausible that the optimal value of these variables reaches zero due to the construction of the consecutive stages, especially in stage two or whenever the vehicle reaches the end of a series of corridors. In the vicinity of the terminal state (in the final corridor) it is typically impractical to use time-optimal control as all stage timings will tend to go to zero and the optimal control problem encounters numerical instability. To avoid this behavior, we actively close one or two of the corridors in these specific cases by fixing this stage time to a small positive value.

\subsection{Optimization Toolchain}
The optimal control problem is formulated easily in Python using the Rockit toolbox, presented in \cite{rockit}, and solved with Ipopt using ma27 as linear solver (\cite{Waechter2005}, \cite{hsl}). The computations are performed on an AMD\textsuperscript{\textregistered} Ryzen~7 pro 3700u processor with eight cores at 2.3~GHz and 29.4~GiB of RAM.

\subsection{MPC Update Strategy}
The multi-stage optimal motion planning problem is solved using a modified MPC approach to account for variable computation times along the trajectory. The update strategy sets the initial condition of the OCP with future state information originating from the previous optimal trajectory, which is, assuming good trajectory tracking performance, a good estimate of the future vehicle state. The size of the time window up to this future state is a trade-off between being sufficiently large to ensure that the MPC is updated before the vehicle reaches the corresponding point, and being sufficiently short to allow trajectory adjustments close to the current vehicle position. 

Due to these on-trajectory updates, the local planner does not incorporate feedback on the vehicle state. A linear feedback controller is implemented to keep the vehicle close to the optimal trajectory, in order to improve tracking performance and to reject disturbances. When a new solution becomes available, it is stitched to the previous solution at the update point from which the initial condition was taken. This approach avoids discontinuous jumps in the reference trajectory for the feedback controller.

\subsection{Linear Stabilizing Feedback Control}
\begin{figure}
    \centering
    \includegraphics[width=0.9\linewidth]{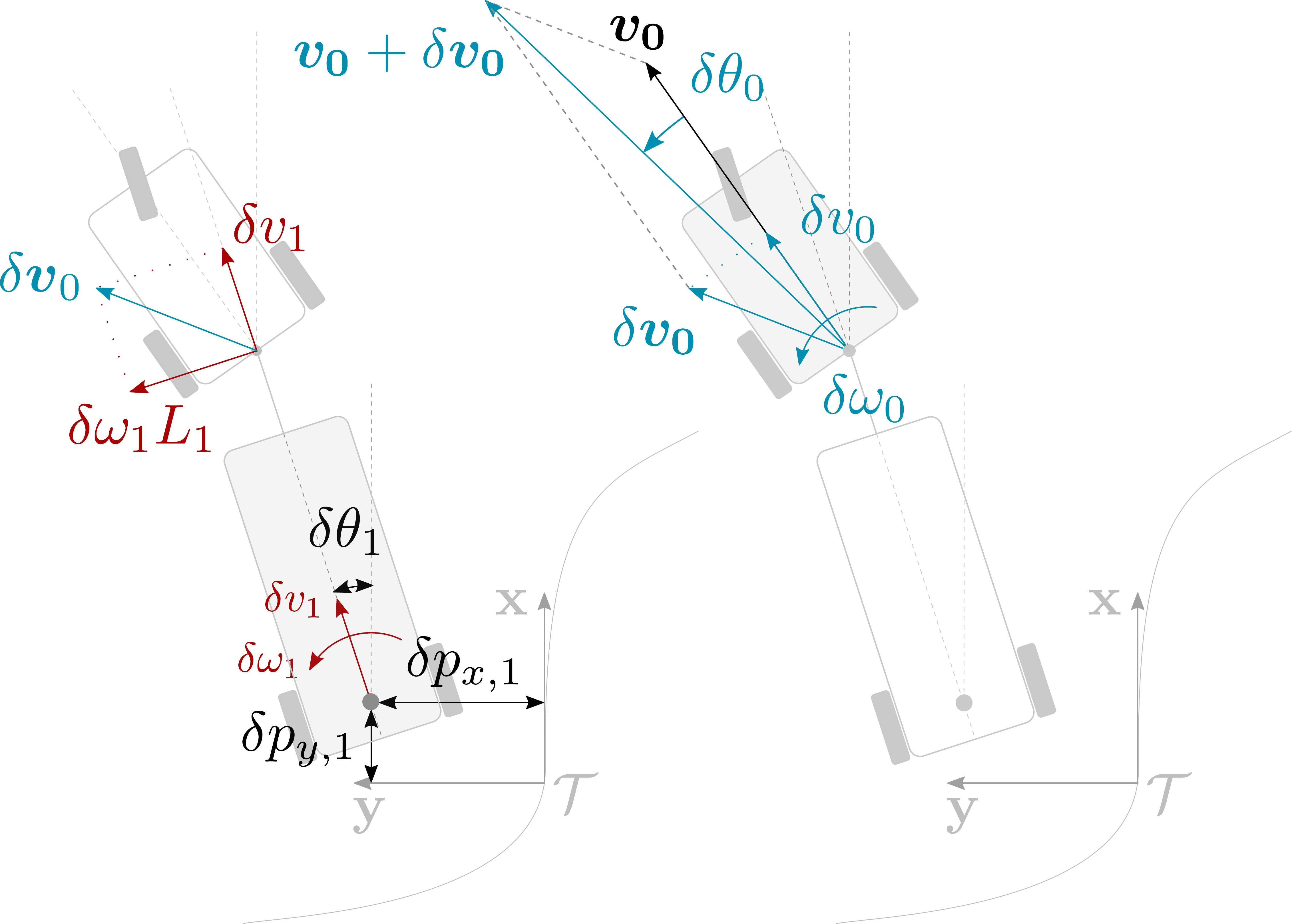}
    \caption{Cascaded state feedback control for stabilization and tracking.}
    \label{fig:fb_ctrl}
\end{figure}

The feedback control scheme for stabilization and tracking is a cascade of two state feedback controllers, one for each vehicle, where the trailer controller generates a reference for the truck controller based on its deviation from the feedforward trajectory. The cascade is set up as follows.

The position and orientation error of the trailer axle's center point, defined as $\delta\vector{x}_1 = [\delta p_{x,1} \: \delta p_{y,1} \: \delta \theta_{1}]\transp$ is expressed in the trajectory coordinate frame~$\mathcal{T}$, which has its $\mathbf{x}$ and $\mathbf{y}$ axes tangential and normal to the feedforward trajectory, as shown in Fig.~\ref{fig:fb_ctrl}. Expressing the error in $\mathcal{T}$ ensures that both the position error $\delta\vector{p}_1$ and the orientation errors of both vehicles $\delta\theta_i$ can be assumed to be small.

The feedback law for the trailer is based on the one presented by \cite{jacobs_agv} for trajectory tracking of a single AMR. It has the shape
\begin{equation}
\begin{aligned}
	\delta\vector{u}_1 &= \begin{bmatrix} \delta v_{1} \\ \delta \omega_1 \end{bmatrix}\\
					 &= -\begin{bmatrix} K_{x,1} & 0 & 0\\ 0 & K_{y,1} & K_{\theta,1} \end{bmatrix}\begin{bmatrix} \delta p_{x,1}\\ \delta p_{y,1}\\ \delta\theta_{1} \end{bmatrix},
\end{aligned}
\end{equation}

where $\delta v_{1}$ and $\delta\omega_1$ are respectively the (small) longitudinal and angular velocity correction of the trailer and $K_{x,1}$, $K_{y,1}$ and $K_{\theta,1}$ are linear feedback gains. \cite{jacobs_agv} present a design procedure to compute linear parameter varying (LPV) feedback gains with vehicle velocity as parameter. Here, we simply make the feedback gains piecewise linear functions of the vehicle velocity, with one slope for forward and one for backward driving, and a dead zone without corrections around the uncontrollable zero velocity point.

The desired trailer corrective action $\delta\vector{u}_1$ is transformed to a reference for the truck as illustrated on the left in Fig.~\ref{fig:fb_ctrl}. The longitudinal and angular velocities at the trailer axle's center point are transformed to desired longitudinal and perpendicular velocities of the truck hitching point of magnitudes $\delta v_1$ and $\delta\omega_1 L_1$, added together into $\delta\vector{v}_0$. For the vehicle used in our experiments, the feedback controller neglects the small $M_0$.

The desired truck velocity correction $\delta\vector{v}_0$ is transformed to the control input correction $\delta\vector{u} = [\delta v_0 \: \delta\omega_0]\transp$, which is added to the feedforward action $\vector{u}$ as on the right in Fig.~\ref{fig:fb_ctrl}. $\delta v_0$ is the projection of $\delta\vector{v}_0$ onto the longitudinal direction of the truck. $\delta\omega_0$ is taken proportional to the orientation difference $\delta\theta_0$ between the feedforward truck velocity $\vector{v}_0$ and the desired truck velocity $\vector{v}_0 + \delta\vector{v}_0$. 

\subsection{Software Framework for Fast Testing and Deploying}
The developed control structure is implemented in a Robot Operating System (ROS) framework, which naturally handles network communication between onboard and offboard software. The onboard deployed software is the interface for the actuated wheel and the onboard sensor interface. The offboard deployed software is a finite-state machine for discrete switching between tasks, the motion planning algorithm, the controller, the localization interface, and visualization tools.

The software configuration with independent ROS nodes additionally entails the advantage that the physical vehicle with actuator and sensor interfaces can easily be replaced by an equivalent simulation node, without any changes in the remainder of the framework. This interchangeability makes the setup ideal for fast offline testing and debugging using the simulator, and direct deployment on the experiment setup afterwards. A vehicle dynamics simulator is easily extracted from the OCP formulation in Rockit.
\section{Experimental Results and Discussion}
\label{experiments}
\begin{figure}
    \centering
    \includegraphics[width=0.8\linewidth]{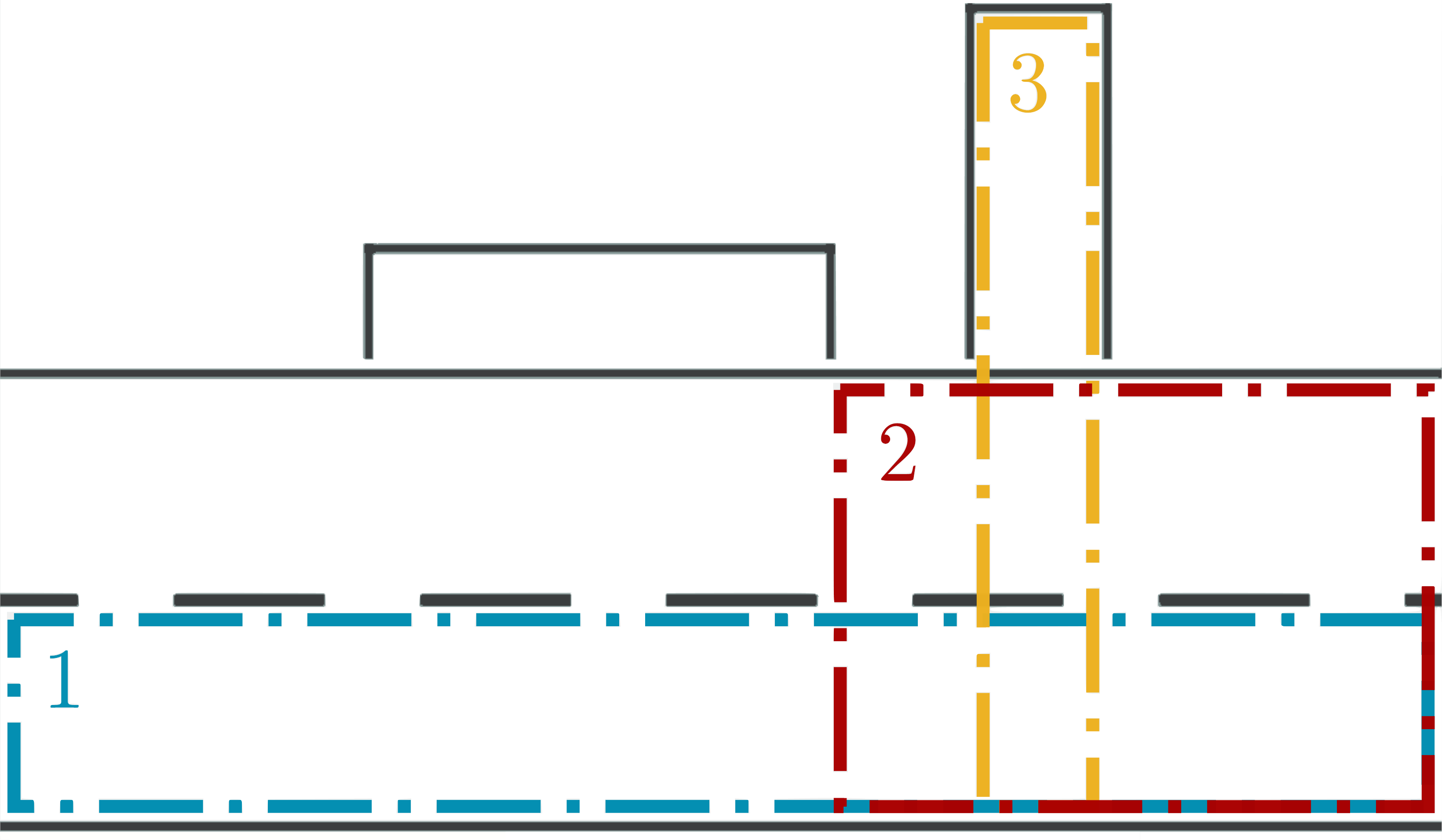}
    \caption{Corridors to travel through during the perpendicular parking maneuver.}
    \label{fig:corridors_road_case}
\end{figure}

The presented approach is validated on a case study of two parking maneuvers, i.e. parallel and perpendicular backward parking. Fig.~\ref{fig:corridors_road_case} shows the three corridors that are subsequently used as box constraints in the multi-stage OCP formulation for the first parking maneuver. First, the AMR drives on the right lane in corridor~1. Next, when it approaches the driveway, it can use the full width of the road in corridor~2. Finally, on the driveway it is restricted to corridor~3. A video of this experiment can be found at \url{https://youtu.be/UKiRny89fzw}. After presenting the lab experiment setup, this section discusses the tracking error, and the importance of initialization of the OCP.

\subsection{Lab Experiment Setup}
\label{section:experiment_lab_setup}

The truck-trailer AMR is the in-house developed test setup shown in Fig.~\ref{fig:truck-trailer_AGV}. It is driven by a KELO-wheel from \cite{kelo}, which takes as input commands a desired rotation rate of the vehicle that it is attached to, and a desired longitudinal velocity, which explains the choice of the control input vector $\vector{u}$ in Section~\ref{subsection:model}. High-level commands from the motion planning algorithm and feedback controller, and onboard sensor measurements are processed by an Odroid~XU4.

\begin{figure}
    \centering
    \includegraphics[width=\linewidth]{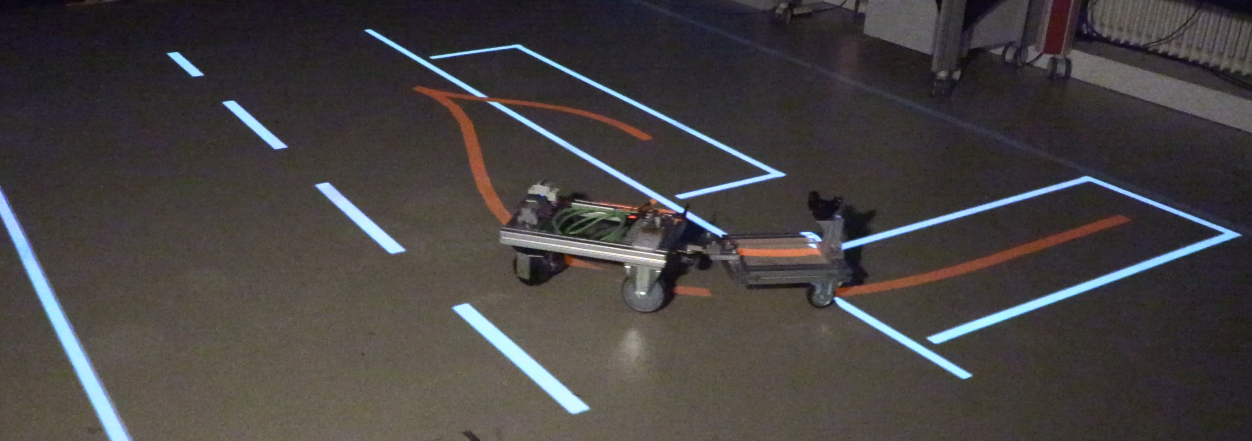}
    \caption{Lab setup with the truck-trailer AMR and projected environment with planned trajectory.}
    \label{fig:lab_setup}
\end{figure}

The vehicle is localized by two infrared emitting lighthouses and a tracker of the HTC Vive virtual reality gaming console as in \cite{vive}, which provides 6D pose estimates at a rate of 100~Hz. The tracker is mounted above the trailer axle's center point, and directly measures states $p_{x,1}$, $p_{y,1}$ and $\theta_1$. An angle encoder is mounted on the hitching point of the truck and measures $\beta_{01}$. Together, these sensors measure the full state $\vector{x}$.

The environment is visualized on the lab floor by two projectors mounted on the ceiling. Apart from a (road) map, occupancy grid, obstacles, or other environmental features, the ROS visualization tool Rviz can also display planned and traveled trajectories, velocity vectors, and other useful debugging information, which makes this a visually attractive setup that is ideal for development and demonstration purposes. The lab setup during the experiment performed for this paper is shown in Fig.~\ref{fig:lab_setup}.

\subsection{Results and Discussion}
\begin{figure}
    \centering
    \begin{minipage}{0.49\linewidth}
        \includegraphics[width=\linewidth]{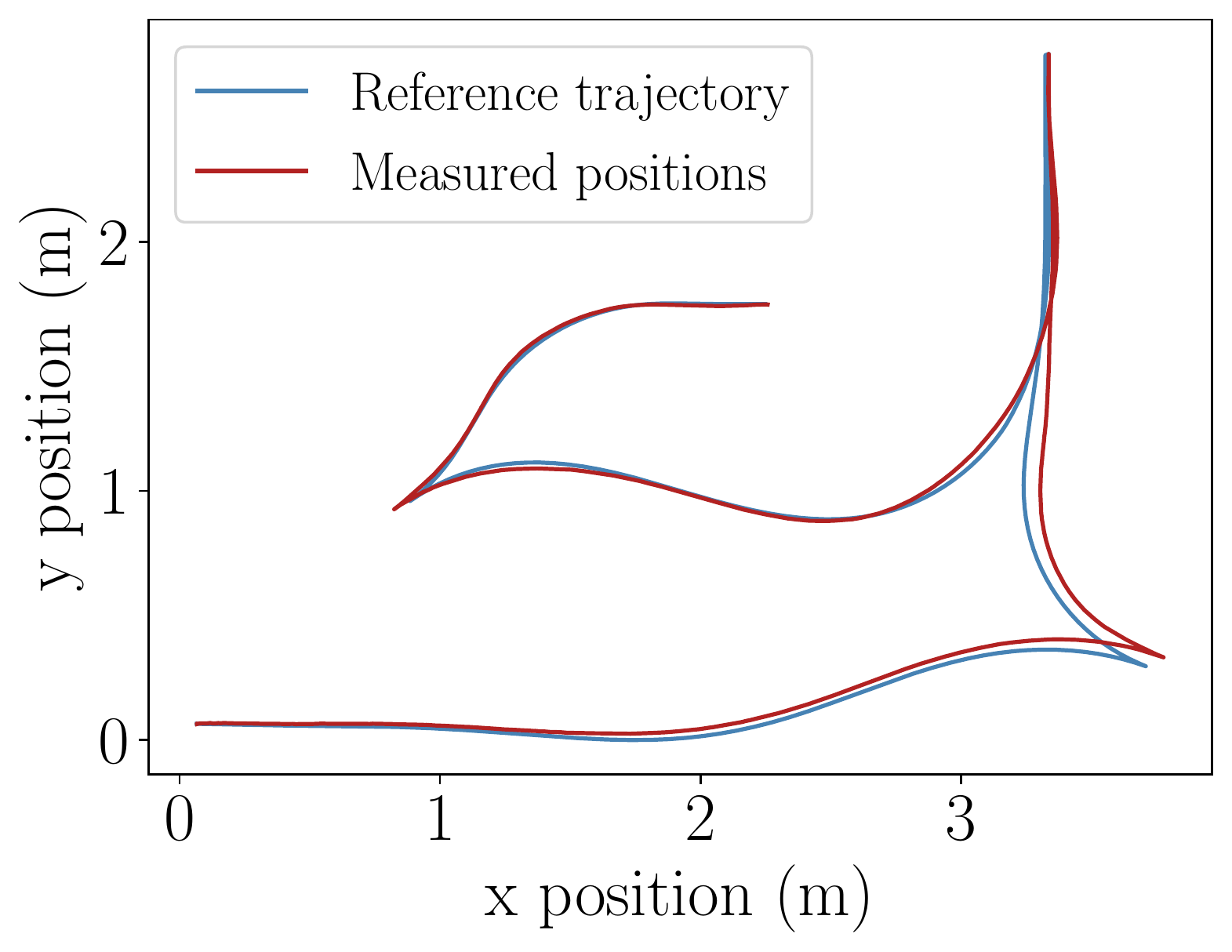}
        \subcaption{Optimal trajectory.}
    \end{minipage}
    \begin{minipage}{0.49\linewidth}
        \includegraphics[width=\linewidth]{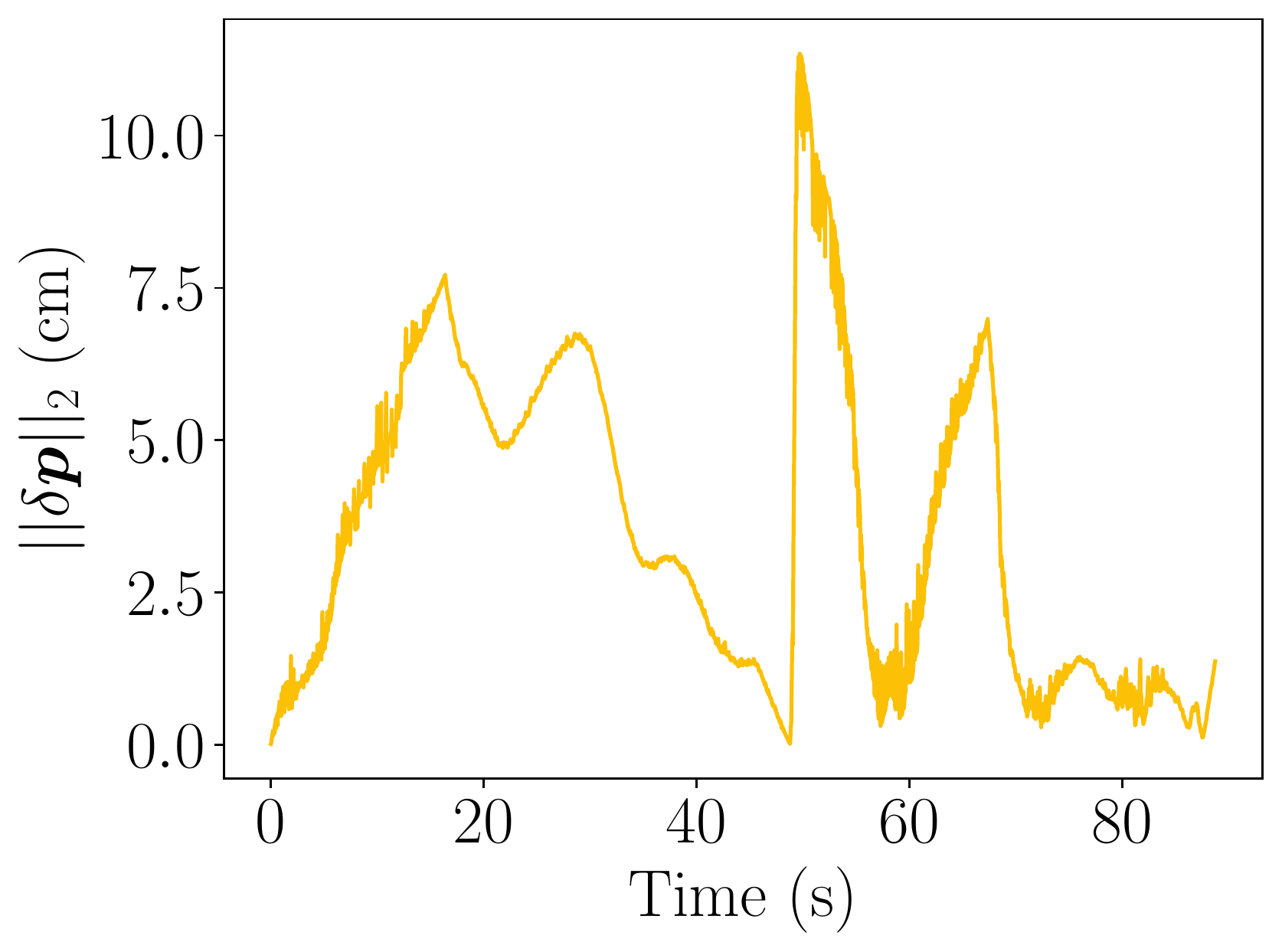}
        \subcaption{Tracking Error.}
     \end{minipage}
     \caption{Trailer reference positions and measured positions (a), and difference between the reference and the measured positions (b) for the full experiment.}
    \label{fig:path}
\end{figure}

Firstly, the performance of the presented approach is evaluated experimentally as described in the previous section. Fig.~\ref{fig:path} shows both the optimal trajectory as well as the traveled path and the tracking error along the trajectory for the lab experiment of the parallel and perpendicular backward parking case. The limited tracking error shows the adequate performance of both the motion planner and feedback controller. Note that the sudden rise in tracking error at the start of the second part of the experiment at 49~s is due to the low-level control implementation of the KELO wheel: at that point, the wheel has its front side towards the back of the vehicle because it just finished parking backward, and when it takes off again out of the parking spot, the wheel first has to make a half turn. While turning, it already applies the desired forward velocity, which results in both a lag in the longitudinal direction and a lateral error, due to the temporarily unmatched direction of the wheel during the rotation. This problem would not occur on an actual truck that can actuate its wheels both in the forward and the backward direction.

\begin{figure}
    \centering
    \includegraphics[width=0.9\linewidth]{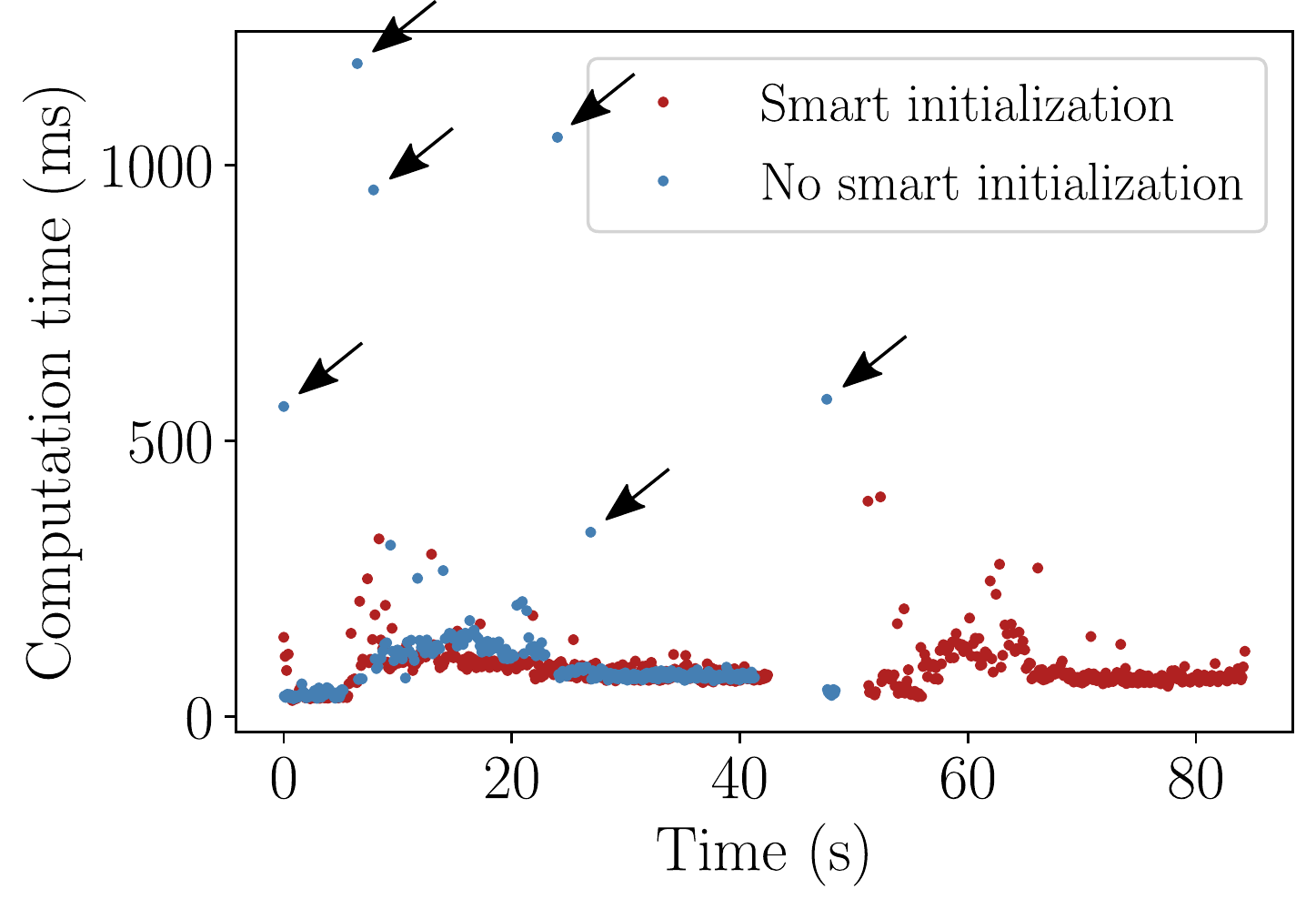}
    \caption{Computation times in simulation with and without smart initialization at the corridor changes. Arrows indicate high computation times when entering a new corridor without smart initialization.}
    \label{fig:calc_time}
\end{figure}

Secondly, Fig.~\ref{fig:calc_time} shows the importance of properly initializing the OCP. Every data point indicates the computation time of an MPC update during two simulation runs of the experiment: one run with smart initialization and one without smart initialization. After the first parking maneuver, around 45~s, there is a short break as the finite-state machine waits for the command to continue the experiment.

This smart initialization is handled as follows. Whenever a solution is available, we can provide this solution as an initial guess for the next iteration. This solution includes all state and control variables together with stage timings and slack variables. At distinct points in time, i.e. at the start of the experiment and whenever a new corridor is entered by one of the vehicles, no initial guess from a previous OCP solution is available. At these points, only some of the optimization variables can be initialized with a close to feasible sequence of states and controls. The state sequence is built by linearly interpolating between the current initial and feasible target points of the active corridors. The control sequence is built considering no rotation and a maximum speed in the expected driving direction.

If this initialization is omitted, there is significant uncertainty that the solver is either not able to find a solution at all, takes a long time to converge, or converges to a spurious local minimum. The computation times, with and without initialization at the change of corridors, are shown in red and blue respectively. At the start of the second part of the experiment, the solver was not able to find a solution within the given time frame when the initialization was omitted which explains the missing data points. Providing a proper initialization allows to find a robust optimal trajectory in a reasonable time window.

\section{Conclusion}
\label{conclusion}

This paper presents a multi-stage optimal control approach to maneuvering of a truck-trailer AMR in a complex environment. The approach consists of two contributions. Firstly, it extends the idea of dividing the environment into convex corridors to these geometrically and kinematically complex vehicles. Secondly, it proposes a strategy to combine computationally expensive Model Predictive Control for trajectory planning with fast linear feedback control that avoids discontinuous jumps in the planned trajectory and control signals to perform online planning updates with stable and smooth forward and backward driving. 

To validate the approach, a lab setup with projected environment and planning visualizations demonstrates a reverse parking use case. The modular software implementation structure allows for fast development in simulation and direct deployment on the experiment setup.

Future work includes the introduction of time-varying corridors, which allows for online environment estimation to adapt for uncertainty due to measurement noise and environment changes or moving obstacles. This addition to the current implementation will show the added value of online re-planning which is already present in the current work. 


\makeatletter
\newcommand*\mysize{%
  \@setfontsize\mysize{9.5}{9.5}%
}
\makeatother
\renewcommand*{\bibfont}{\mysize}

\bibliography{paper}             

\end{document}